%% file: acl_latex.tex
\documentclass[11pt]{article}

\usepackage[final]{acl}

\usepackage{times}
\usepackage{latexsym}

\usepackage[T1]{fontenc}

\usepackage[utf8]{inputenc}

\usepackage{microtype}

\usepackage{inconsolata}

\usepackage{graphicx}

\usepackage{amsmath}
\usepackage{amsfonts}
\usepackage{amssymb}
\usepackage{bbding}
\usepackage{booktabs}
\usepackage{dsfont}
\usepackage{multicol}
\usepackage{multirow}
\usepackage{rotating}
\usepackage{tipa}
\usepackage{wasysym}
\usepackage{times}
\usepackage{helvet}
\usepackage{courier}
\usepackage{xcolor}
\usepackage{url}
\usepackage{pifont}

\usepackage{CJKutf8}
\usepackage{makecell}
\usepackage[most]{tcolorbox}    

\newtcolorbox{promptbox}[1]{
    fontupper=\scriptsize,
    colback=gray!5!white,    
    colframe=gray!75!black,  
    fonttitle=\bfseries,     
    title={#1},              
    boxrule=0.5pt,           
    left=5pt, right=5pt, top=5pt, bottom=5pt
}

\title{Aligning Language Models with Real-time Knowledge Editing}

\author{Chenming Tang$^*$ \quad
Yutong Yang$^*$ \quad
Kexue Wang \quad
Yunfang Wu$^\dag$ \\
  National Key Laboratory for Multimedia Information Processing, Peking University \\
  School of Computer Science, Peking University \\
  \small{\texttt{\{tangchenming, 2300013219\}@stu.pku.edu.cn}}\quad\quad \small{\texttt{\{yytpku, wuyf\}@pku.edu.cn}}\\
  \small{$^*$Equal contribution \quad $^\dag$Corresponding author}
}

\begin{document}
\maketitle

\begin{abstract}
Knowledge editing aims to modify outdated knowledge in language models efficiently while retaining their original capabilities. Mainstream datasets for knowledge editing are predominantly static and fail to keep in pace with the evolving real-world knowledge. In this work, we introduce CRAFT, an ever-evolving real-world dataset for knowledge editing. It evaluates models on temporal locality, common-sense locality, composite portability and alias portability, providing a comprehensive and challenging evaluation for knowledge editing, on which previous methods hardly achieve balanced performance. Towards flexible real-time knowledge editing, we propose KEDAS, a novel paradigm of knowledge editing alignment featuring diverse edit augmentation and self-adaptive post-alignment inference, exhibiting significant performance gain on both CRAFT and traditional datasets compared to previous methods. We hope this work may serve as a catalyst for shifting the focus of knowledge editing from static update to dynamic evolution.\footnote{All of the code and data of this work are publicly available at~\url{https://github.com/JamyDon/CRAFT-KEDAS}.}
\end{abstract}

\section{Introduction}

Once language models (LMs) are pre-trained, knowledge is injected into them and becomes their static internal capability~\cite{LLM-as-knowledge-base}. As time goes by, some knowledge in LMs inevitably becomes incorrect or out of date, and it is vital to update the outdated knowledge. However, re-training a model frequently is highly costly and usually infeasible. To this end, the techniques of knowledge editing have been developed~\cite{survey} to edit specific knowledge in LMs efficiently without heavy re-training.

There have been a large number of datasets to evaluate the effectiveness of knowledge editing approaches. However, most traditional ones like ZsRE~\cite{zsre}, MQuAKE~\cite{mquake} and RippleEdits~\cite{ripple} are static and not real-time. Once the datasets are constructed and released, they remain fixed and can not be updated anymore, which is far from real-world knowledge editing applications. Recently, there have been a few real-time datasets released. For example, WikiBigEdit~\cite{wikibigedit} collects real-world changes in Wikipedia, but it requires processing massive Wiki data (hundreds of gigabytes) from the Internet, which is costly and  inconvenient in practice. Meanwhile, the automatically collected dataset lacks necessary filtering and thus suffers from severe sparsity (for instance, only 3.6\% of the collected data have an entry of portability evaluation). EvoWiki~\cite{evowiki} is another real-time dataset based on Wikipedia, but it focuses on retrieval-augmented generation (RAG) and continual learning and does not support comprehensive evaluation for knowledge editing due to lack of crucial aspects like locality. Overall, there still lacks a high-quality, easily-updated dataset for real-world and real-time knowledge editing.

\input{tables/benchmark}

In this work, we introduce CRAFT (\textbf{\underline{C}}hinese \textbf{\underline{R}}eal-time statistics \textbf{\underline{A}}nd \textbf{\underline{F}}inance knowledge editing datase\textbf{\underline{T}}), an ever-evolving real-world dataset for knowledge editing in Chinese. It leverages publicly available official data that are continuously updated, ensuring both temporal freshness and real-world applications. CRAFT is organized in well-designed paired edits, each serving as a composite reasoning test to evaluate models' ability to integrate multiple related factual updates. Meanwhile, it supports evaluations on alias portability and temporal and common-sense locality, providing a comprehensive assessment of LMs' adaptability and robustness under dynamic knowledge changes. The differences between CRAFT and previous datasets are summarized in Table~\ref{tab:benchmark_comparison}.

We perform exposure analysis on CRAFT and other evaluation datasets with five different LMs. The results indicate that a large portion of knowledge in existing datasets has been known to LMs while the exposure rate of CRAFT is nearly zero. This validates the real-time property of CRAFT, which makes it challenging and fair.


We evaluate a suite of representative knowledge editing methods on CRAFT and demonstrate current approaches exhibit inherent limitations. Parameter-based approaches like ROME~\cite{rome} and WISE~\cite{wise} suffer from gradual model degradation and scarcely edit successfully when new knowledge comes sequentially and continually. Simple retrieval-based approaches like IKE~\cite{ike} and EREN~\cite{eren} fail to achieve consistent balanced performance due to lack of alignment. LTE~\cite{lte}, which aligns LMs with knowledge editing via finetuning, shows weak locality due to overfitting on irrelevant queries.

Towards flexible real-time knowledge editing, we propose KEDAS\footnote{\textipa{/"ki:d@s/}, pronounced as ``kee-das''.}, namely \textbf{\underline{K}}nowledge \textbf{\underline{E}}diting alignment with \textbf{\underline{D}}iverse \textbf{\underline{A}}ugmentation and \textbf{\underline{S}}elf-adaptive inference, an advanced knowledge editing framework featuring diverse representations of edits and flexible inference paths. The one-time offline alignment stage first finetunes the LM with low-rank adaptation (LoRA)~\cite{lora}. Once the alignment is finished, users may edit any knowledge without modifying any parameters. During editing, we introduce diverse edit augmentation to store different forms of each edit in the memory. In the inference phase, a filter-enhanced smart retriever is employed to adaptively select the base model or the aligned model. If any edits are recalled, the prompt will go through the LM with LoRA activated. Otherwise, the prompt just passes to the original LM to guarantee locality and avoid over-fitting. On both CRAFT and traditional datasets, KEDAS exhibits outstanding performance in all metrics and significantly outperforms previous methods, illustrating an ideal paradigm of knowledge editing alignment.

Our contributions are as follows:
\begin{itemize}
    \item We reveal the leakage issue of existing datasets and introduce CRAFT, an open-source, contamination-free and easily-updated dataset for real-time knowledge editing.
    \item We evaluate existing knowledge editing methods on CRAFT and reveal the limitations of these methods in real-time knowledge editing, especially the tradeoff among edit success, locality and portability.
    \item We propose KEDAS, a novel paradigm of aligning LMs with real-time knowledge editing which significantly outperforms existing methods with balanced performance.
\end{itemize}

This paper is organized as follows. We introduce the preliminary of knowledge editing in \S~\ref{sec:preliminary}. Then, we present our CRAFT dataset in \S~\ref{sec:craft} and propose KEDAS for real-time knowledge editing in \S~\ref{sec:kedas}. After that, we conduct empirical experiments in \S~\ref{sec:experiments}. Finally, we discuss related work in \S~\ref{sec:related-work} and conclude our study in \S~\ref{sec:conclusion}.

\section{Preliminary}\label{sec:preliminary}
\subsection{Task Formulation}\label{subsec:formulation}
An LM system $f$ can be regarded as a function $f : \mathcal{Q} \mapsto \mathcal{A}$ mapping an query $q\in\mathcal{Q}$ to its output answer $a = f(q)\in\mathcal{A}$.

Knowledge editing aims to change the behavior of the LM after modifying some knowledge. Edits of knowledge are usually in the form of query-answer (QA) pairs:
\begin{equation}
\mathcal{E} = \{e^t\}^{N}_{t=1} = \{(q^t_e, a^t_e)\}^{N}_{t=1},
\end{equation}
where $q^t_e$ is an input query triggering knowledge (\textit{e.g.}, \texttt{The current US president is}), $a^t_e$ is the corresponding target of editing (\textit{e.g.}, \texttt{Donald Trump}), $t$ is the index of each edit, and $N$ denotes the total number of edits.

To assess the efficacy of knowledge editing, the post-edit LM $f^*$ is evaluated via the following criteria~\cite{survey}:

\paragraph{Edit success}
measures the accuracy of editing, requiring $f^*$ to correctly recall the edits:
\begin{equation}
\mathbb{E}_{(q_e, a_e)\in\mathcal{E}}\ \mathds{1}\{f^*(q_e) = a_e\}.
\end{equation}
\paragraph{Locality}
measures the precision of editing, requiring $f^*$ not to change its behavior out of the scope of edits:
\begin{equation}
\mathbb{E}_{(q_l, a_l)\in\mathcal{E}_l}\ \mathds{1}\{f^*(q_l) = f(q_l)\},
\end{equation}
where $\mathcal{E}_{l}$ are QA pairs of unrelated queries. Note that the target of locality is unchanged answer rather than exactly the ground truth answer.
\paragraph{Portability}
measures how well edits transfer to related queries, requiring $f^*$ to correctly answer such queries:
\begin{equation}
\mathbb{E}_{(q_p, a_p)\in\mathcal{E}_p}\ \mathds{1}\{f^*(q_p) = a_p\},
\end{equation}
where $\mathcal{E}_{p}$ are QA pairs related to existing edits.

\subsection{Settings of Knowledge Editing}\label{subsec:ke-setting}

\paragraph{Single Editing}
In single editing, each edit is evaluated directly after it is applied to the original model $f$: 

\begin{equation}
f^t = \mathbf{Edit}(f, q^t_e, a^t_e),\quad1\le t\le N.
\end{equation}

This traditional setting is widely employed by earlier work. It is far from real-world applications, since only one edit can be applied.

\paragraph{Sequential Editing}
In sequential editing, edits are applied step by step in a continual manner ($f^0$ denotes $f$ for consistency):
\begin{equation}
\label{eq:seq-edit}
f^t = \mathbf{Edit}(f^{t-1}, q^t_e, a^t_e),\quad1\le t\le N.
\end{equation}
The final model $f^N$ after applying all $N$ edits is evaluated. This setting is frequently used to evaluate the lifelong performance and scalability of knowledge editing. We focus on this setting in this work because it is closer to practical applications, especially the real-time knowledge editing setting.

\subsection{Knowledge Editing Alignment}
\label{subsec:lte}

LTE~\cite{lte} relates knowledge editing to LM alignment, aligning LMs with ever-changing real-time knowledge edits. During alignment, LMs’ capabilities of knowledge updating are enlightened by an editing prompt:
\begin{verbatim}
    [Updated Information] {edit}
    [Query] {query}
\end{verbatim}
The training data consist of in-scope and out-of-scope queries with or without the specific edit to improve edit efficacy while keeping locality. 

In the inference phase, LMs are required to conduct on-the-fly and streaming knowledge editing by retrieving relevant edits to the query from the stored memory.

\input{tables/dataset_stats}

\section{The CRAFT Dataset}\label{sec:craft}
We introduce CRAFT, a real-time and real-world dataset designed to evaluate knowledge editing under dynamically evolving factual contexts. Unlike static datasets that rely on outdated or widely exposed Wikipedia facts, CRAFT continuously updates with publicly available official data sources, ensuring fairness and real-world relevance. 

\subsection{Data Sources}

CRAFT consists of two complementary subsets, namely \textbf{CRAFT-Statistics} and \textbf{CRAFT-Finance}, representing two highly correlated real-world domains that evolve periodically. Example instances from the two subsets are shown in Appendix~\ref{app:stat} and \ref{app:fin} respectively.

\paragraph{CRAFT-Statistics}
We collect monthly statistical reports from the National Bureau of Statistics of China\footnote{\url{https://data.stats.gov.cn/}}, covering July~2024 to June~2025 in the current version used in this work. The dataset spans four major domains -- finance, fiscal affairs, telecommunications, and transportation -- containing 221 indicators and 4,468 data points. Our data collection pipeline, leveraging the \texttt{cn-stats} API\footnote{\url{https://github.com/songjian/cnstats}}, supports automatic temporal updates, enabling effortless collection of the most recent monthly data.

\paragraph{CRAFT-Finance}
CRAFT-Finance includes annual financial statements of 390 publicly listed Chinese companies (2023–2024), collected via the \texttt{AKShare} API\footnote{\url{https://github.com/akfamily/akshare}}. It features financial indicators such as debt ratio, shareholder equity, and operating cash flow, supporting yearly updates to reflect the latest financial disclosures.

\subsection{Evaluation Metrics}
\label{subsec:craft-metrics}

To comprehensively assess model adaptability, CRAFT provides four evaluation dimensions besides the conventional edit success (ES) as described in \S~\ref{subsec:formulation}:

\paragraph{Composite Reasoning Portability (P$_{\text{reasoning}}$)}
We propose \emph{composite reasoning portability}, a new evaluation designed specifically for CRAFT. 
Unlike traditional \emph{multi-hop reasoning portability}, which constructs reasoning chains from a single edit, 
composite portability evaluates a model's ability to integrate multiple distinct edits into a single reasoning query. 

In CRAFT, all data are organized in \emph{paired format}: each test instance consists of \emph{two edits} and one corresponding composite reasoning query. Formally, given a pair of edits $\{e_1, e_2\}$, the composite portability query $q_p$ requires reasoning over both edits simultaneously. For example, consider a CRAFT-Finance instance with a pair of edits:
\begin{align*}
    e_1 &: \text{2024 Company A's total assets = 2M} \\
    e_2 &: \text{2024 Company A's total liabilities = 1M} 
\end{align*}
The corresponding composite reasoning query and answer are:
\begin{align*}
    q_p &: \text{2024 Company A's debt ratio} \\
    a_p &: 0.5
\end{align*}
CRAFT-Finance includes 10 types of such combinations, demonstrating the generality and practical relevance of composite reasoning portability. For CRAFT-Statistics, composite portability is constructed by pairing two edits of the same indicator from adjacent years.

\paragraph{Alias Portability (P$_\text{alias}$)}  
We propose \emph{alias portability} to evaluate a model's robustness to synonymous indicator names by replacing key terms with domain-specific aliases constructed from a manually curated alias list.

\paragraph{Temporal Locality (L$_\text{temporal}$)}
We propose \emph{temporal locality} to how well a model's unrelated general knowledge and reasoning capability are preserved after knowledge editing.

\paragraph{Common-sense Locality (L$_\text{common}$)}
We propose \emph{common-sense locality} to measure whether a model's irrelevant general knowledge and reasoning capability remain consistent after knowledge editing. 
For CRAFT, the questions for this evaluation are directly sampled from the C$^\text{3}$ dataset~\cite{sun2019investigating}, which contains 19,577 Chinese multiple-choice comprehension questions.

\subsection{Exposure Analysis}

A perfect evaluation dataset should be unleaked, with a low portion of knowledge has been exposed to LMs (\textit{i.e.}, a low exposure rate). If a large portion of knowledge has already been known by LMs, the evaluation results on it can be inaccurate, unfair, or even misleading.

We evaluate the exposure rates of CRAFT and existing benchmarks across five representative LMs by directly prompting with the questions in the datasets and counting how many questions can be correctly answered. As shown in Table~\ref{tab:dataset_stats}, Due to its temporal freshness and domain specificity, CRAFT exhibits extremely low exposure rates (nearly 0\%) across all the five LMs, mitigating data contamination and ensuring fair evaluation. Meanwhile, for other datasets, a great portion of the knowledge has been known to the LMs, making them less challenging and less fair in practice. This reveals the advantage of CRAFT as a truly real-time dataset.

\subsection{Dataset Statistics}

\input{tables/Craft_stat}

Table~\ref{tab:craft_dense_stats} shows the numbers of instances in CRAFT's subsets. Composite reasoning portability is evaluated per edit pair, and other metrics per edit. For experiments, we split each subset of CRAFT into training and test sets. The test set contains 500 instances from the two subsets each (1,000 instances in total) and the training set contains the remaining instances.

\subsection{Discussion}

CRAFT is a scalable and continuously evolving dataset reflecting real-world knowledge updates. 
Its dynamic nature, factual authenticity, and comprehensive reasoning evaluation make it a robust foundation for future research on reliable knowledge editing. Anyone who wants to construct CRAFT within an arbitrary time period can easily obtain a customized dataset with our provided open-source scripts. Through CRAFT, we also demonstrate that researchers can make full use of public available data sources (like public APIs) besides the widely used Wikipedia data to create datasets with real-world and real-time properties and scalability that can help evaluate, train or develop more scalable intelligent systems without much human labor.

\section{KEDAS: A Proposal for Real-time Knowledge Editing Alignment}\label{sec:kedas}

\input{figures/KEDAS}

As will be shown in the experiments (\S~\ref{sec:experiments}), previous methods show certain limitations either in recall (\textit{i.e.}, edit success and portability) or precision (\textit{i.e.}, locality) on CRAFT. To improve edit success and generalization while keeping a good locality in real-time knowledge editing, we propose KEDAS, an all-round approach consisting of several complementary designs.

The overall framework of KEDAS is illustrated in Figure~\ref{fig:KEDAS}, including alignment, editing and inference. In alignment, the LM is aligned with knowledge editing using LoRA. In editing, each edit is converted into diverse forms and stored to the memory module. In the inference phase, edits are retrieved from the memory based on the user query and then filtered by a trained classifier. If there exist relevant edits, the model with trained LoRA adapters will be called for inference. Otherwise, the query will go through the frozen base model. The order of editing and inference is flexible based on practical needs (``edit as you go''), as shown by the bidirectional arrow. For example, we may apply several edits and employ the LM for inference immediately. If there come further new edits, we can return to the editing phase to perform these edits and then resume inference.

\subsection{Alignment Based on LoRA}
LTE~\cite{lte} suffers from a limited locality because it employs the post-alignment model, whose behavior can be significantly different from the original one, to all incoming queries. To address this, we propose the idea of self-adaptive post-alignment inference. To achieve this elegantly, we align the base model via parameter-efficient finetuning with LoRA~\cite{lora}, which tunes additional adapters during alignment. This costs low computational resources for both training and inference, making KEDAS efficient. We follow LTE to construct alignment data using the training set of CRAFT and details are in Appendix~\ref{app:algn-data}.

\subsection{Diverse Edit Augmentation}
\input{tables/edit_aug}

In previous work, each piece of edited knowledge is directly applied to the system as it is, limiting generalization. Since each knowledge can be expressed in various ways, leveraging just one fixed form is inflexible, especially for challenging portability evaluations like that in CRAFT.

In this work, we propose diverse edit augmentation, a novel technique that augments each edit by converting it into multiple expressions. The \textit{de facto} standard form of edits in knowledge editing is a QA pair, typically describing the relation between entities or attributes of them. Based on this observation, we manually design several ways of form augmentation, including \textit{declarative} and \textit{aliased} forms, as demonstrated in Table~\ref{tab:edit-aug}.

Concretely, for an edit $e^t = (q^t_e, a^t_e)$ at step $t$\footnote{For simplicity, we deem the two edits in each CRAFT instance as two separate time steps.}, its conventional QA form is:
\begin{equation}
    e^t_\text{qa} = q^t_e \oplus a^t_e,
\end{equation}
where $\oplus$ denotes concatenation. Then, it is augmented as:
\begin{align}
&e^t_\text{dec} = \mathbf{Declarative}(q^t_e, a^t_e),\\
&e^t_\text{als} = \mathbf{Aliased}(q^t_e, a^t_e).
\end{align}
Specifically, \textit{declarative} is to convert the QA pair into a declaration while \textit{aliased} is to replace specialized terms in the declaration with their aliases for more flexible retrieval. In experiments, we conduct these augmentations by prompting an API-based off-the-shelf LM.

We store the augmented diverse edits into memory for future retrieval:
\begin{equation}
    \mathcal{M}^0 = \emptyset,\quad
    \mathcal{M}^t = \mathcal{M}^{t-1} \cup \{e^t_\text{qa}, e^t_\text{dec}, e^t_\text{als}\},
\end{equation}
where $\mathcal{M}^t$ denotes the memory at time step $t$.

Note that our augmentation methods are specifically designed for the CRAFT dataset. There can be more diverse augmentations like paraphrasing in practice for broader use.

\subsection{Filter-based Smart Retriever}
During inference, the common practice regarding retrieval in previous knowledge editing work is to directly employ an embedder and perform single-stage retrieval to obtain the top-$k$ hits. This suffers from a poor locality due to the lack of a filter measuring the relevance of queries. In this work, we propose a filter-based smart retriever that pairs the high-recall retrieval with a high-precision filter by precisely identifying relevant queries. In this way, on challenging datasets like CRAFT, the post-edit system can keep consistent behaviors for neighboring locality queries and preserve its original answers to general queries.

First, we retrieve top-$n$ candidates from the memory using a normal embedding-based retriever. Thanks to our diverse edit augmentation, the memory contains various forms of edits, improving the recall rate. At the time step $t$, for an input query $q$, a set of top-$n$ candidates $\mathcal{C}^{\text{top-}n}$ is retrieved as:

\begin{equation}
    \mathcal{C}^{\text{top-}n} = \mathbf{Retriever}^{\text{top-}n}(q, \mathcal{M}^t).
\end{equation}

Then, a binary classifier $\theta$ is exploited to  predict the relevance between candidates and a query:
\begin{equation}
    \mathcal{C}^\text{filtered} = \{e\mid \mathbf{Filter}_\theta(e) = 1, e\in \mathcal{C}^{\text{top-}n}\}.
\end{equation}
Then we only adopt the top-$k$ edits with the highest similarity to the query if the filtered candidate set contains too many edits:
\begin{equation}
\mathcal{E}^* =\operatorname{arg\,top}_k(\mathbf{Sim}(e, q)) \ \text{for } e \in \mathcal{C}^\text{filtered}.
\end{equation}

\subsection{Self-adaptive Post-alignment Inference}
In the inference phase, we employ both the original pre-alignment model and the post-alignment adapter, which does not consume extra memory while enabling two inference paths. Each time a query $q$ is requested, the filter-based smart retriever is utilized to recall possible relevant edits in the memory. If there exists a set of relevant edits $\mathcal{E}^*$, indicating that the query may involve edited knowledge, we fill both $\mathcal{E}^*$ and $q$ into the knowledge editing prompt (see \S~\ref{subsec:lte}) and then pass the prompt to the model with the post-alignment adapters activated. Otherwise, the query directly goes through the pre-alignment model without any processing. The inference strategy is formularized below:

\small
\begin{equation}
    a = \begin{cases}
    f_{\Phi}(q) & \text{irrelevant}\\
    f_{\Phi + \Delta\Phi(\Theta)}(\mathbf{KEPrompt}(\mathcal{E}^*, q)) & \text{relevant}\\
    \end{cases},
\end{equation}
\normalsize
where $a$ denotes the final answer, $f_\Phi$ denotes the original model parametrized by $\Phi$, $f_{\Phi + \Delta\Phi(\Theta)}$ denotes the model with post-alignment adapters $\Delta\Phi(\Theta)$, and $\mathbf{KEPrompt}(\cdot, \cdot)$ denotes the knowledge editing prompt template. 

\subsection{Discussion}
KEDAS borrows the idea of inference-time retrieval from EREN~\cite{eren} and the idea of knowledge editing alignment from LTE~\cite{lte}. The novelty of KEDAS lies in the two unique designs: diverse edit augmentation and filter-based smart retriever paired with self-adaptive post-alignment inference. The former follows the idea that edits can go beyond their initial forms, while the latter demonstrates the insight that we do not have to perform knowledge editing for irrelevant queries, and a lightweight mechanism leads to more balanced performance.

\input{tables/main}

\section{Experiments}\label{sec:experiments}


\subsection{Experimental Setup}
\subsubsection{Evaluation}
We evaluate representative knowledge editing methods and our proposed KEDAS with CRAFT. We use the metrics described in \S~\ref{subsec:craft-metrics}, including ES, L$_\text{temporal}$, L$_\text{common}$, P$_\text{reasoning}$ and P$_\text{alias}$.

We also validate the effectiveness of KEDAS in traditional knowledge editing with four widely used datasets from KnowEdit~\cite{survey}, including ZsRE~\cite{zsre}, WikiBio~\cite{wikibio}, WikiData$_\text{recent}$ and WikiData$_\text{counterfact}$~\cite{wikidata}. We use the metrics described in \S~\ref{subsec:formulation} for evaluation. See Appendix~\ref{app:knowedit_data} for details.

In the main experiments, we focus on sequential editing for both evaluation datasets.

\subsubsection{Language Models}
For CRAFT, we use Llama-3.1-8B-Instruct~\cite{llama3} and Qwen2.5-3B-Instruct~\cite{qwen25}, both from widely used herds of open-source LMs (\textsc{Llama-3.1} and \textsc{Qwen2.5}) and with different scales. For KnowEdit, we focus on the same \textsc{Llama-3.1} model and provide the results on two other LMs in Appendix~\ref{app:knowedit}.

\subsubsection{Evaluated Methods}

We evaluate the following knowledge editing methods:
(1)~\textbf{LoRA}~\cite{survey} directly updates knowledge by LoRA-based finetuning.
(2)~\textbf{ROME}~\cite{rome} locates multilayer perceptron weights of related knowledge and writes in new key-value pairs.
(3)~\textbf{IKE}~\cite{ike} leverages in-context examples of copying, updating and retaining knowledge to edit knowledge via prompting.
(4)~\textbf{EREN}~\cite{eren} prompts LMs to decide the relevance of retrieved edits and then queries LMs with or without edits.
(5)~\textbf{WISE}~\cite{wise} stores edits in a parametric side memory module and utilized a router to decide which memory to go through.
(6)~\textbf{LTE}~\cite{lte} aligns LMs with the knowledge editing task by post-training.
(7)~\textbf{KEDAS} is our proposed method.

Due to compatibility, cost, and performance issues, we do not include some other baselines as discussed in Appendix~\ref{app:baselines}.

\subsubsection{Implementation Details}
Our experiments are based on the framework of EasyEdit~\cite{easyedit}. For most of the baselines, we follow the default hyper-parameter settings provided by EasyEdit.

For IKE, we only evaluate the 8-shot setting because of limited resources. For both LTE and KEDAS, we use paraphrase-multilingual-MiniLM-L12-v2~\cite{sbert} as the retriever and set the number of adopted edits $k$ as 3, the default setting of LTE. For KEDAS, we adopt bert-base-chinese~\cite{bert} as the binary filter, which is trained based on the training set of CRAFT with details specified in Appendix~\ref{app:bert-data}. We set the number of retrieved candidates $n$ as 8 and that of adopted edits $k$ as 3, in line with LTE. The diverse edit augmentation of KEDAS is done by prompting \texttt{gpt-4o-mini}\footnote{\url{https://platform.openai.com}} and the prompts are in Appendix~\ref{app:dea}.

\subsection{Results on CRAFT}

The results on CRAFT are presented in Table~\ref{tab:main}.

Previous methods of knowledge editing show limitations in certain aspects. LoRA and ROME, as traditional methods primarily designed for single editing or a limited number of edits, show no advantage when applied to the sequential editing setting on CRAFT. IKE, the retrieval-based approach, suffers from poor locality and mediocre portability, albeit achieves outstanding edit success. The prompting-based EREN requires specific prompt optimization for different LMs and can scarcely perform editing with its default prompt, as can be seen from the low edit success. WISE is parameter-based and suffers from downgrading performance under the sequential setting, with weak performance in most metrics. LTE, with an alignment stage, achieves relatively robust performance in edit success and portability but struggles to achieve locality due to its fixed inference path. These indicate that CRAFT is challenging for existing knowledge editing methods, making it hard to achieve a consistently balanced performance.

Meanwhile, KEDAS exhibits outstanding performance on CRAFT, securing the highest or second-highest scores in all the metrics. The edit success of KEDAS is nearly 100\% while all of its locality scores are 100\% except for L$_\text{temporal}$ on the Finance subset possibly because the BERT-based filter suffers from limited generalization in certain cases. Meanwhile, the portability of KEDAS is significantly outstanding on \textsc{Llama-3.1} and is better or on par with LTE on \textsc{Qwen2.5}. Overall, KEDAS outperforms WISE, the representative parameter-based approach, by \textbf{+35.63} and \textbf{+34.73} averaged points on the two LMs respectively, and surpasses LTE, the strong alignment-based baseline, by \textbf{+36.52} and \textbf{+32.30} averaged points. This confirms that KEDAS achieves the most balanced performance on CRAFT and maintains competitive performance across all dimensions.

\subsection{Results on KnowEdit}\label{subsec:knowedit}
\input{tables/knowedit}

The results on KnowEdit are shown in Table~\ref{tab:knowedit}. KEDAS exhibits outstanding performance in all the metrics and secures the highest harmonic mean (HM) scores across all the datasets. On average, its HM outperforms WISE by \textbf{+26.9} and surpasses LTE by \textbf{+18.2}. As an alignment-based approach, KEDAS is superior especially in locality compared with LTE, with an improvement of \textbf{+31.7}.

On KnowEdit, we also further experiment with other LMs and under the traditional setting of single editing. These results can be found in Appendix~\ref{app:knowedit}, where KEDAS also secures consistent advantage with balanced performance.

\subsection{Ablation Study}

\input{tables/ablations}

In this section, we assess the indispensability of components of KEDAS, including diverse edit augmentation (DEA), filter (FLT) and self-adaptive post-alignment inference (SPI). The results are presented in Table~\ref{tab:ablations}. The exclusion of diverse edit augmentation leads to a decline in edit success and portability, indicating that it contributes to the recall of the retriever. Meanwhile, both the filter and the paradigm of self-adaptive post-alignment inference show great contribution to locality, demonstrating that they are playing important roles in KEDAS. These results confirm that all the core components of KEDAS make contributions to the final performance and are indispensable.

\section{Related Work}\label{sec:related-work}

\subsection{Related Evaluation Datasets}

Conventionally, knowledge editing datasets are created based on temporal changes or counter-fact modifications of world knowledge either from existing QA datasets~\cite{zsre, survey-2023} or Wikipedia~\cite{ripple, mquake}, in which \citet{mquake} first propose multi-hop composite portability. However, these are all static and suffer from data contamination as time goes by.

\citet{wikibigedit} propose WikiBigEdit, a dynamic dataset based on periodical changes in Wikipedia. However, only a small fraction (3.6\%) of its samples support reasoning portability evaluation, limiting its comprehensiveness. Similarly, \citet{evowiki} propose EvoWiki, a Wiki-based real-time dataset primarily designed for RAG. Although EvoWiki claims to automatically update with new edits, its implementation and data pipeline remain unpublished.

Besides conventional knowledge editing datasets, there have also been some designed for special use. \citet{serac} propose ConvSent to edit LMs' sentiments on given topics.~\citet{sanitation} propose Sanitation to forget specific information stored in LMs and thus address privacy concerns. \citet{hallueditbench} propose HalluEditBench to evaluate whether knowledge editing correct hallucinations.

Our proposed CRAFT leverages statistical and financial data, which are both real-time and inherently dynamic. The strong interdependence among indicators allows each pair of edits to correspond to a meaningful reasoning portability test. This enables comprehensive evaluation across five dimensions: data type, real-time property, real-world grounding, reasoning portability, and composite reasoning portability.

\subsection{Knowledge Editing Methods}

\paragraph{Parameter-based Editing}
This line of work edits model parameters or adds extra parameters for knowledge editing. \citet{rome}, \citet{memit} and \citet{alphaedit} adopt a locate-then-edit manner and aim to precisely edit MLP weights. \citet{mend} adopt meta-learning~\cite{meta-learning} to predict the change of model parameters.~\citet{calinet}, \citet{t-patcher} and \citet{wise} add extra parameters to store edits in a parametric form. \citet{serac} train a separate counter-fact model and does not involve retrieval.

\paragraph{Retrieval-based Editing}
This line of work stores edits in a memory module and retrieve from it when needed. \citet{ike} leverage in-context examples of copying, updating and retaining knowledge to edit knowledge via prompting. \citet{eren} first prompt LMs to decide the relevance of retrieved edits and then query LMs with or without in-context edits, which may seem conceptually related to KEDAS but is essentially different in that it relies merely on prompting. \citet{recipe} convert edits into soft prompt tokens with trained models and then concatenate user prompts to modify the behavior of frozen LMs.~\citet{lte} align LMs with knowledge editing by post-training with in-context edits. Our proposed KEDAS can also be categorized as retrieval-based.

\section{Conclusion}\label{sec:conclusion}
We have introduced CRAFT, a novel automatically updatable real-time dataset for knowledge editing that focuses on national statistics and finance of China. We evaluate representative knowledge editing methods on CRAFT and reveal that existing methods have certain limitations. We then propose a simple yet effective approach, KEDAS, to better align LMs with real-time knowledge editing. Experiments validate the effectiveness of KEDAS on CRAFT in all aspects including edit success, locality and portability. We hope this work may serve as a catalyst for shifting the focus of knowledge editing from static update to dynamic evolution.

\section*{Acknowledgments}
This work is supported by the National Natural
Science Foundation of China (62076008).

\section*{Limitations}
\paragraph{Domain} CRAFT focuses on the Chinese language only and is limited to the domains of national statistics and finance. Nevertheless, our methodology is universal and can be applied to any other languages and CRAFT is easily expandable to domains that data are periodically updated and automatically accessible.
\paragraph{Portability} The composite reasoning portability of CRAFT is merely 2-hop. Nevertheless, it can be expanded to multiple hops via combination of time and indicators.
\paragraph{Baselines} Only a limited set of existing methods is evaluated in our experiments due to limited resources and compatibility issues.
\paragraph{Filter} The performance of KEDAS heavily depends on the filter. If the filter is not sufficiently robust, the overall performance of KEDAS will also be less competitive.

\section*{Ethical Considerations}
\paragraph{Computational Budget}
All our experiments are conducted on a machine with Ubuntu 20.04.6 LTS, Intel$^\circledR$ Xeon$^\circledR$ Silver 4310 CPU and 256G memory. We use one NVIDIA A40 48G GPU for all the experiments. The training of KEDAS takes about 2 hours. The editing of KEDAS takes about 50 minutes on both subsets of CRAFT.

\paragraph{Reproducibility}
Our work is fully reproducible since we have provided our implementation details and scripts for experiments.

\paragraph{Potential Risks}
To the best of our knowledge, there are no potential risks concerning our work.

\paragraph{Scientific Artifacts}
\input{tables/licenses}
We cite all the creators of scientific artifacts we use in this paper. Licenses of these scientific artifacts are shown in Table~\ref{tab:license}. Our use of these artifacts is consistent with their intended use.

\paragraph{Privacy and Offense Concerns}
Our data source ensures that our CRAFT dataset contains no information that names or uniquely identifies individual people or offensive content.


\bibliography{custom}

\clearpage

\appendix

\section{Details of Alignment Data for KEDAS}
\label{app:algn-data}
For each instance of CRAFT's training set $\mathcal{E}^t$, we construct the edit candidate set $\mathcal{E}^*$ by keeping the two edits $e_1$ and $e_2$ of the instance and adding 0-3 (randomly decided) extra retrieved edits with the same retriever as LTE that are different from the two to promote model's robustness to incorrect retrieved results.

To promote edit success, for each edit query $q_e$ and answer $a_e$ of either $e_1$ or $e_2$, the input is $\mathbf{KEPrompt}(\mathcal{E}^*, q_e)$ and the target is $a_e$.

To promote portability, for each portability query $q_p$ and answer $a_p$ in $\mathcal{E}^t_p$, the input is $\mathbf{KEPrompt}(\mathcal{E}^*, q_p)$ and the target is $a_p$.

To promote locality, for each locality query $q_l$ and answer $a_l$ in $\mathcal{E}^t_l$, the in-scope input is $\mathbf{KEPrompt}(\mathcal{E}^*, q_l)$ with the target $a_l$. We also include an out-of-scope input $q_l$ with the target $a_l$ that is without any retrieved edits.

In this way, the LM learns to leverage retrieved edits for relevant queries while keeping the answer unchanged for locality queries.

\section{Details of LM Alignment}
\label{app:LM-algn}

\input{tables/hparams}

We adopt LLaMA-Factory~\cite{llamafactory} to finetune the LM. The alignment is done on an NVIDIA A40 48GB GPU and takes 2 hours.

The main training hyperparameters are presented in Table~\ref{tab:hparams}, most remain unchanged as the default values. Note that we train the LM for only one epoch because finetuning for multiple epoches will cause over-fitting.

\input{figures/promp_declarative}

\section{Details of Diverse Edit Augmentation}
\label{app:dea}

The prompt templates of diverse edit augmentation for \texttt{gpt-4o-mini} are presented in Figure~\ref{fig:prompt-declarative} and~\ref{fig:prompt-aliased}. We include manually written in-context examples in the templates to better instruct the model.

\section{Details of Training Data for the Filter}
\label{app:bert-data}
For each instance of CRAFT's training set $\mathcal{E}^t$ with $\mathcal{E}^t_e$, $\mathcal{E}^t_p$ and $\mathcal{E}^t_l$ as the set of edits, portability QAs and locality QAs respectively, we simply construct the training data for the filter based on relevance. For any two edits $e=(q_e, a_e), e'=(q_{e'}, a_{e'})$ (may be identical) in $\mathcal{E}^t_e$, the input is $q_{e'} \texttt{[sep]} e$ and the target is $1$ (relevant). For any edit $e\in\mathcal{E}^t_e$ and any portability QA pair $(q_p, a_p)\in\mathcal{E}^t_p$, the input is $q_p \texttt{[sep]} e$ and the target is $1$ (relevant). For any edit $e\in\mathcal{E}^t_e$ and any locality QA pair $(q_l, a_l)\in\mathcal{E}^t_l$, the input is $q_l \texttt{[sep]} e$ and the target is $0$ (irrelevant). In this way, the filter learns to decide the relevance between the query and the edit.

\section{Details Regarding KnowEdit}\label{app:knowedit_data}
\input{tables/knowedit_stat}
Table~\ref{tab:knowedit-stat} presents the data statistics of KnowEdit.

Following~\citet{survey} and~\citet{lte}, our evaluation metrics for KnowEdit include edit success (\textbf{ES}), locality (\textbf{L}) and portability (\textbf{P}). We also calculate the harmonic mean (\textbf{HM}) of the three scores as the edit quality. Note that WikiBio does not contain portability queries and we thus calculate the harmonic mean based on the first two metrics.

\section{Other Baselines}\label{app:baselines}
In our primary experiments, we also tried to evaluate many other modern baselines but encountered different sorts of issues (including compatibility, availability, cost, performance, \textit{etc.}).

When running MEMIT~\cite{memit}, our GPU ran out of memory. Meanwhile, it performed poorly in all the metrics on some datasets (\textit{e.g.}, merely 1.91 and 0.75 edit success on WikiData$_\text{recent}$ and WikiData$_\text{counterfact}$. Since MEMIT is primarily designed for single editing, we attribute the poor performance to its incompatibility with the sequential editing setting.

AlphaEdit~\cite{alphaedit}, a more advanced method designed on the basis of MEMIT, also exhibited relatively poor performance (for instance, its edit success scores on WikiData$_\text{recent}$ and WikiData$_\text{counterfact}$ were merely 20.60 and 5.91 respectively) so that we do not include it as a baseline for comparison. We believe this is due to its incompatibility with the sequential editing setting or the LMs we use for experiments.

MEND~\cite{mend}, SERAC~\cite{serac} and RECIPE~\cite{recipe} require a lot of resources that we cannot afford currently for training and do not provide trained checkpoints for the \textsc{Llama-3} herd of LMs used in our experiments. Therefore, we do not include these in our main experiments.

Nevertheless, for a comprehensive comparison, we experiment on KnowEdit with all the mentioned baselines under the single editing setting using Llama-2-7B-Chat~\cite{llama2} due to that all the methods are compatible with single editing and the \textsc{Llama-2} herd of LMs. The results are presented in Appendix~\ref{app:knowedit}.

\input{figures/prompt_aliased}

\section{Example from CRAFT-Statistics}
\label{app:stat}

\begin{CJK}{UTF8}{gbsn}

\begin{itemize}
  \item \textbf{Prompt 1:} 2024年7月中国的货币和准货币(M2)供应量期末值(亿元)是多少？  
  \textit{(English translation: What is the period-end value (100 million yuan) of China's Money and Quasi-Money (M2) supply in July 2024?)}  
  \textbf{Target New:} 3033060.78

  \item \textbf{Prompt 2:} 2023年7月中国的货币和准货币(M2)供应量期末值(亿元)是多少？  
  \textit{(English translation: What is the period-end value (100 million yuan) of China's Money and Quasi-Money (M2) supply in July 2023?)}  
  \textbf{Target New:} 2854031.56
\end{itemize}

\vspace{0.5em}
\noindent\textbf{Portability}

\begin{itemize}
  \item \textbf{Alias:}
  \begin{itemize}
    \item Prompt: 2024年7月中国的Money and Quasi-Money (M2) Supply, period-end(100 million yuan)是多少？  
    \textit{(English translation: What is the Money and Quasi-Money (M2) Supply, period-end (100 million yuan), in July 2024 in China?)}  
    Answer: 3033060.78

    \item Prompt: 2023年7月中国的Money and Quasi-Money (M2) Supply, period-end(100 million yuan)是多少？  
    \textit{(English translation: What is the Money and Quasi-Money (M2) Supply, period-end (100 million yuan), in July 2023 in China?)}  
    Answer: 2854031.56
  \end{itemize}

  \item \textbf{Composite Reasoning:}  
  Prompt: 2024年7月中国的货币和准货币(M2)供应量期末值(亿元)比2023年7月的高多少？  
  \textit{(English translation: How much higher was China's M2 supply (100 million yuan) in July 2024 than in July 2023?)}  
  Answer: 179029.22
\end{itemize}

\vspace{0.5em}
\noindent\textbf{Locality}

\begin{itemize}
  \item \textbf{Temporal:}
  \begin{itemize}
    \item Prompt: 2022年7月中国的货币和准货币(M2)供应量期末值(亿元)是？  
    \textit{(English translation: What was the M2 supply (100 million yuan) at the end of July 2022 in China?)}
  
    Answer: 2578078.57

    \item Prompt: 2025年7月中国的货币和准货币(M2)供应量期末值(亿元)是？  
    \textit{(English translation: What is the M2 supply (100 million yuan) at the end of July 2025 in China?)}  
    Answer: 3299429.06
  \end{itemize}

  \item \textbf{Common-sense:}
  \begin{itemize}
    \item Prompt: 请根据以下材料回答问题。  
    （略）问题：动物的器官感觉与人的相比有什么不同？  
    \textit{(English translation: Read the following passage and answer the question. Question: How do animals’ sensory organs differ from those of humans?)}  
    Answer: D

    \item Prompt: 请根据以下材料回答问题。  
    （略）问题：录音中提到能预报风暴的动物是什么？  
    \textit{(English translation: Read the following passage and answer the question. Question: Which animal mentioned in the passage can predict storms?)}  
    Answer: C
  \end{itemize}
\end{itemize}

\end{CJK}

\section{Example from CRAFT-Finance}
\label{app:fin}

\begin{CJK}{UTF8}{gbsn}

\begin{itemize}
  \item \textbf{Prompt 1:} 2024年三友医疗的总资产（亿元）是多少？  
  \textit{(English translation: What is the total assets (100 million yuan) of Sanyou Medical in 2024?)}  
  \textbf{Target New:} 23.07

  \item \textbf{Prompt 2:} 2024年三友医疗的总负债（亿元）是多少？  
  \textit{(English translation: What is the total liabilities (100 million yuan) of Sanyou Medical in 2024?)}  
  \textbf{Target New:} 2.58
\end{itemize}

\vspace{0.5em}
\noindent\textbf{Portability}

\begin{itemize}
  \item \textbf{Alias:}
  \begin{itemize}
    \item Prompt: 2024年三友医疗的Total Assets(100 million yuan)是多少？  
    \textit{(English translation: What is the total assets (100 million yuan) of Sanyou Medical in 2024?)}  
    Answer: 23.07

    \item Prompt: 2024年三友医疗的Total Liabilities(100 million yuan)是多少？  
    \textit{(English translation: What is the total liabilities (100 million yuan) of Sanyou Medical in 2024?)}  
    Answer: 2.58
  \end{itemize}

  \item \textbf{Composite Reasoning:}  
  Prompt: 2024年三友医疗的资产负债率（\%）是多少？  
  \textit{(English translation: What is the debt-to-asset ratio (\%) of Sanyou Medical in 2024?)}  

  Answer: 11.2
\end{itemize}

\vspace{0.5em}
\noindent\textbf{Locality}

\begin{itemize}
  \item \textbf{Temporal:}
  \begin{itemize}
    \item Prompt: 2023年三友医疗的总资产（亿元）是多少？  
    \textit{(English translation: What was the total assets (100 million yuan) of Sanyou Medical in 2023?)}  
    Answer: 22.61

    \item Prompt: 2023年三友医疗的总负债（亿元）是多少？  
    \textit{(English translation: What was the total liabilities (100 million yuan) of Sanyou Medical in 2023?)}  
    Answer: 2.19
  \end{itemize}

  \item \textbf{Common-sense:}
  \begin{itemize}
    \item Prompt: 请根据以下材料回答问题。  
    （略）问题：小羊的结局是什么？  
    \textit{(English translation: Read the following passage and answer the question. Question: What happened to the little lamb?)}  
    Answer: C

    \item Prompt: 请根据以下材料回答问题。  
    （略）问题：这个故事告诉我们什么道理？  
    \textit{(English translation: Read the following passage and answer the question. Question: What moral lesson does this story convey?)}  
    Answer: A
  \end{itemize}
\end{itemize}

\end{CJK}

\section{More Results on KnowEdit}
\label{app:knowedit}

\input{tables/knowedit_all}

The sequential editing results with Llama-2-7B-Chat~\cite{llama2} and Qwen2.5-7B-Instruct~\cite{qwen25} are presented in Table~\ref{tab:knowedit-all}. KEDAS secures the highest harmonic mean scores on all the four datasets, outperforming previous methods significantly.

\input{tables/knowedit_single}

We also experiment with Llama-2-7B-Chat under the traditional single editing setting and the results are presented in Table~\ref{tab:knowedit-single}. KEDAS exhibits the most balanced performance among all the 11 evaluated methods.

\end{document}

%% file: tables/benchmark.tex
\begin{table}[t]
\centering
\footnotesize
\begin{tabular}{lcccc}
\toprule
\textbf{Dataset} &
\textbf{\shortstack{RT}} &
\textbf{\shortstack{RW}} &
\textbf{\shortstack{RP}} &
\textbf{\shortstack{CRP}} \\
\midrule
ZsRE & \ding{55} & \checkmark & \ding{55} & \ding{55} \\
MQuAKE-T & \ding{55} & \checkmark & \checkmark & \ding{55} \\
MQuAKE-CF & \ding{55} & \ding{55} & \checkmark & \checkmark \\
RippleEdits & \ding{55} & \checkmark & \checkmark & \ding{55} \\
WikiBigEdit & \checkmark & \checkmark & \ding{55} & \ding{55} \\
EvoWiki & $\triangle$ & \checkmark & \checkmark & \ding{55} \\
CRAFT (Ours) & \checkmark & \checkmark & \checkmark & \checkmark \\
\bottomrule
\end{tabular}
\caption{
Comparison of knowledge editing datasets.  
\textbf{RT} refers to ``real-time" and \textbf{RW} denotes ``real-world".
\textbf{RP} (reasoning portability) tests reasoning chains derived from a single edit,
whereas \textbf{CRP} (composite reasoning portability) evaluates a model’s ability to integrate multiple distinct edits into a single reasoning query.
}
\label{tab:benchmark_comparison}
\end{table}

%% file: tables/dataset_stats.tex
\begin{table*}[t]
\centering
\small
\setlength{\tabcolsep}{1.5pt}
\resizebox{\textwidth}{!}{
\begin{tabular}{lccccccc}
\toprule
\multirow{2}{*}{\textbf{Dataset}} & \textbf{GPT-3.5} & \textbf{Mistral-7B} & \textbf{Llama-3.1-8B} & \textbf{DeepSeek-chat} & \textbf{Doubao-1.5-pro-32k} & \multirow{2}{*}{\textbf{Size}} \\
& \cite{openai2023gpt35} & \cite{jiang2023mistral7b} & \cite{llama3} & \cite{deepseekai2025deepseekv3technicalreport} & \cite{doubao2025} &  \\
\midrule
ZsRE~\cite{zsre}  & 43.98\% & 23.48\% & 22.45\% & 52.47\% & 26.77\% & 6,000 \\
MQuAKE-T~\cite{mquake} & 35.46\% & 76.21\% & 46.11\% & 42.58\% & 76.09\% & 5,604 \\
EvoWiki$_\text{evolved}$~\cite{evowiki} & 8.07\% & 16.22\% & 6.43\% & 10.24\% & 5.08\% & 1,338 \\
EvoWiki$_\text{stable}$~\cite{evowiki} & 28.45\% & 42.37\% & 25.90\% & 39.76\% & 12.32\% & 1,494 \\
EvoWiki$_\text{uncharted}$~\cite{evowiki}  & 8.19\% & 22.62\% & 7.39\% & 16.02\% & 3.43\% & 1,136 \\
CRAFT-Statistics (Ours)  & 0.01\% & 0.00\% & 0.30\% & 1.80\% & 0.99\% & 4,468 \\
CRAFT-Finance (Ours)  & 0.00\% & 0.00\% & 0.14\% & 0.42\% & 0.00\% & 7,800 \\
\bottomrule
\end{tabular}
}
\caption{
Exposure rate (\%) of different LLMs to benchmark datasets.
Model release years are indicated in citations. 
CRAFT exhibits minimal exposure due to real-time data collection and domain freshness.
}
\label{tab:dataset_stats}
\end{table*}

%% file: tables/Craft_stat.tex
\begin{table}[htbp]

\small
\setlength{\tabcolsep}{4pt}      
\renewcommand{\arraystretch}{0.9}  
\begin{tabular}{p{1.5cm}ccccc}
\toprule
\makecell{\textbf{Subset}} & \textbf{Edit} & $\mathbf{P}_\text{reasoning}$ & $\mathbf{P}_\text{alias}$ & $\mathbf{L}_\text{temporal}$ & $\mathbf{L}_\text{common}$ \\
\midrule
\makecell{CRAFT-\\Statistics} & 4,468 & 2,234 & 4,468 & 4,468 & 4,468 \\
\makecell{CRAFT-\\Finance}    & 7,800 & 3,900 & 7,800 & 7,800 & 7,800 \\
\bottomrule
\end{tabular}
\caption{The numbers of instances in different evaluation types of CRAFT. Composite reasoning uses edit pairs; other metrics are per edit.}
\label{tab:craft_dense_stats}
\end{table}

%% file: figures/KEDAS.tex
\begin{figure*}[htbp]
    \centering
    \includegraphics[width=1\linewidth]{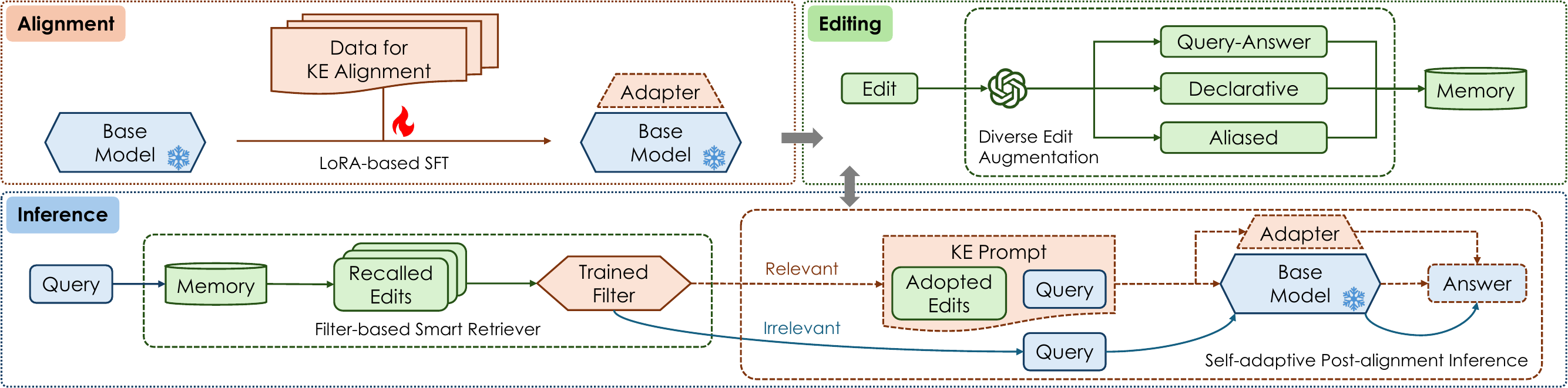}
        \caption{Illustration of our proposed KEDAS framework. }
    \label{fig:KEDAS}
\end{figure*}

%% file: tables/edit_aug.tex
\begin{table}[htbp]
\small
    \begin{tabular}{c|c}
    \toprule
        \textbf{Form} & \textbf{Content} \\
    \midrule
        QA & What are Microsoft's total assets? \$619B.\\
    \midrule
        \textit{Declarative} & Microsoft's total assets are \$619B.\\
        \textit{Aliased} & MSFT's total assets are \$619B.\\
    \bottomrule
    \end{tabular}
    \caption{A demonstration of diverse edit 
    expressions.}
    \label{tab:edit-aug}
\end{table}

%% file: tables/main.tex
\begin{table*}[htbp]
\small
  \centering
  \setlength{\tabcolsep}{1.5mm}
    \begin{tabular}{l|l|ccccc|ccccc}
    \toprule
    \multirow{2}[0]{*}{\textbf{Model}} & \multirow{2}[0]{*}{\textbf{Method}} & \multicolumn{5}{c|}{\textbf{CRAFT-Statistics}}  & \multicolumn{5}{c}{\textbf{CRAFT-Finance}} \\
          & & \textbf{ES} & \textbf{L$_\text{temporal}$} & \textbf{L$_\text{common}$} & \textbf{P$_\text{reasoning}$} & \textbf{P$_\text{alias}$} & \textbf{ES} & \textbf{L$_\text{temporal}$} & \textbf{L$_\text{common}$} & \textbf{P$_\text{reasoning}$} & \textbf{P$_\text{alias}$} \\
    \midrule
    \multirow{7}[0]{*}{\textsc{Llama-3.1}} & LoRA  & 47.93 & 0.71  & 0.00  & 41.56 & 47.62 & 44.19 & 13.97 & 0.00  & 9.81  & 7.43 \\
    & ROME  & 21.59 & 0.00  & 12.20 & 13.67 & 21.59 & 4.25  & 4.00  & 14.00 & 7.62  & 4.40 \\
    & IKE   & \textbf{99.82} & 5.49  & 11.20 & 43.03 & \underline{55.73} & \textbf{99.98} & 27.50 & 11.60 & 40.71 & 57.86 \\
    & EREN  & 22.12 & \underline{91.68} & 63.20 & 24.99 & 22.29 & 24.37 & \textbf{70.93} & 65.80 & 25.48 & 22.27 \\
    & WISE  & 47.54 & 29.74 & \underline{81.00} & 41.26 & 47.30 & 32.66 & 49.62 & \underline{79.80} & 34.41 & 32.48 \\
    & LTE   & 97.90 & 5.44  & 12.60 & \underline{68.58} & 54.14 & 80.68 & 27.94 & 8.40  & \underline{50.21} & \underline{60.98} \\
    & KEDAS & \underline{99.80} & \textbf{100.00} & \textbf{100.00} & \textbf{81.20} & \textbf{62.17} & \underline{99.85} & \underline{62.50} & \textbf{100.00} & \textbf{55.95} & \textbf{70.61} \\
    \midrule\midrule
    \multirow{7}[0]{*}{\textsc{Qwen2.5}} & LoRA  & 43.81 & 18.86 & 0.00  & 39.48 & 35.91 & 24.73 & 11.80 & 0.00  & 31.59 & 32.89 \\
    & ROME  & 35.06 & 32.79 & 0.00  & 29.00 & 26.90 & 9.75  & 2.95  & 0.00  & 14.39 & 9.34 \\
    & IKE   & \underline{96.82} & 19.44 & 82.40 & 48.48 & 42.02 & 97.96 & 28.78 & 81.80 & 39.27 & 46.16 \\
    & EREN  & 28.32 & \underline{61.08} & 92.60 & 34.55 & 30.67 & 31.25 & 42.43 & 93.00 & 33.69 & 29.53 \\
    & WISE  & 40.72 & 43.46 & \underline{94.40} & 34.57 & 38.39 & 28.07 & \underline{51.43} & \underline{95.00} & 28.67 & 21.25 \\
    & LTE   & \textbf{99.95} & 8.38  & 6.60  & \underline{77.67} & \textbf{58.92} & \textbf{99.96} & 31.56 & 5.20  & \textbf{51.86} & \underline{60.18} \\
    & KEDAS & \textbf{99.95} & \textbf{100.00} & \textbf{100.00} & \textbf{80.88} & \underline{58.59} & \underline{99.88} & \textbf{66.47} & \textbf{100.00} & \underline{51.55} & \textbf{65.97} \\
    \bottomrule
    \end{tabular}
  \caption{Main results on CRAFT. The highest and second-highest scores are \textbf{bolded} and \underline{underlined}, respectively.}
  \label{tab:main}
\end{table*}

%% file: tables/knowedit.tex
\begin{table*}[htbp]
\small
\setlength{\tabcolsep}{2mm}
  \centering
    \begin{tabular}{l|cccc|ccc|cccc|cccc}
    \toprule
    {\multirow{2}[0]{*}{\textbf{Method}}} & \multicolumn{4}{c|}{\textbf{ZsRE}} & \multicolumn{3}{c|}{\textbf{WikiBio}} & \multicolumn{4}{c|}{\textbf{WikiData$_\text{recent}$}} & \multicolumn{4}{c}{\textbf{WikiData$_\text{counteract}$}} \\  
     & \textbf{ES} & \textbf{L} & \textbf{P} & \textbf{HM} & \textbf{ES} & \textbf{L} & \textbf{HM} & \textbf{ES} & \textbf{L} & \textbf{P} & \textbf{HM} & \textbf{ES} & \textbf{L} & \textbf{P} & \textbf{HM} \\
    \midrule
    IKE   & 98.5  & 43.5  & 64.4  & 61.7  & 96.9  & 37.9  & 54.5  & 97.4  & 48.8  & 67.2  & 65.7  & 91.3  & 53.7  & 61.4  & 65.4 \\
          EREN  & 32.3  & 70.7  & 44.9  & 44.5  & 59.4  & 64.3  & 61.8  & 44.4  & 73.1  & 37.5  & 47.7  & 23.5  & \textbf{75.4} & 22.9  & 30.2 \\
          WISE  & 60.0  & 51.3  & 40.4  & 49.3  & 84.5  & 95.9  & 89.8  & 68.8  & \textbf{94.5} & 43.1  & 62.1  & 47.1  & 45.4  & 35.0  & 41.8 \\
          LTE   & \textbf{99.6} & 55.5  & 66.7  & 69.7  & \textbf{97.7} & 59.7  & 74.1  & \textbf{99.8} & 52.9  & 74.7  & 70.9  & \textbf{98.3} & 59.3  & 73.4  & 73.8 \\
          KEDAS & \textbf{99.6} & \textbf{90.5} & \textbf{71.4} & \textbf{85.5} & \textbf{97.7} & \textbf{100.0} & \textbf{98.9} & \textbf{99.8} & 73.9  & \textbf{76.1} & \textbf{81.7} & \textbf{98.3} & 69.2  & \textbf{77.4} & \textbf{79.9} \\
    \bottomrule
    \end{tabular}
  \caption{Results on KnowEdit. Metrics include edit success (ES), locality (L), portability (P) and harmonic mean (HM) of the previous metrics. The highest scores are \textbf{bolded}.} 
  \label{tab:knowedit}
\end{table*}

%% file: tables/ablations.tex
\begin{table*}[htbp]
\small
  \centering
    \begin{tabular}{l|ccccc|ccccc}
    \toprule
    \multirow{2}[0]{*}{\textbf{Method}} & \multicolumn{5}{c|}{\textbf{CRAFT-Statistics}}  & \multicolumn{5}{c}{\textbf{CRAFT-Finance}} \\
          & \textbf{ES} & \textbf{L$_\text{temporal}$} & \textbf{L$_\text{common}$} & \textbf{P$_\text{reasoning}$} & \textbf{P$_\text{alias}$} & \textbf{ES} & \textbf{L$_\text{temporal}$} & \textbf{L$_\text{common}$} & \textbf{P$_\text{reasoning}$} & \textbf{P$_\text{alias}$} \\
    \midrule
    KEDAS & \textbf{99.80} & \textbf{100.00} & \textbf{100.00} & \textbf{81.20} & \textbf{62.17} & 99.85 & \textbf{62.50} & \textbf{100.00} & \textbf{55.95} & \textbf{70.61} \\
    \midrule
    \ w/o DEA & 96.51 & \textbf{100.00} & \textbf{100.00} & 71.02 & 59.89 & 98.10 & \textbf{62.50} & \textbf{100.00} & 53.90 & 63.35 \\
    \ w/o FLT & \textbf{99.80} & 5.44  & 11.40 & \textbf{81.20} & \textbf{62.17} & \textbf{100.00} & 27.48 & 8.40  & \textbf{55.95} & \textbf{70.61} \\
    \ w/o SPI & \textbf{99.80} & 5.68  & 10.80 & \textbf{81.20} & \textbf{62.17} & 99.85 & 27.42 & 8.40  & \textbf{55.95} & \textbf{70.61} \\
    \bottomrule
    \end{tabular}
  \caption{Ablation results on CRAFT. The highest scores are \textbf{bolded}.}
  \label{tab:ablations}
\end{table*}

%% file: tables/licenses.tex
\begin{table}[htbp]
\small
  \centering
    \begin{tabular}{ll}
    \toprule
    \textbf{Artifact} & \textbf{License} \\
    \midrule
    C$^\text{3}$ & Non-commercial Purpose\\
    cn-stats & MIT \\
    AKShare & MIT \\
    BERT  & Apache \\
    MiniLM & Apache \\
    EasyEdit \& KnowEdit & MIT \\
    \textsc{Llama-3} & Llama 3 Community License \\
    \textsc{Qwen2.5} & Apache \\
    \textsc{Llama-2} & Llama 2 Community License \\
    \bottomrule
    \end{tabular}%
  \caption{Licenses of scientific artifacts we use.}
  \label{tab:license}%
\end{table}%

%% file: tables/hparams.tex
\begin{table}[htbp]
\small
  \centering
    \begin{tabular}{cc}
    \toprule
    \textbf{Hyperparameter} & \textbf{Value} \\
    \midrule
    Batch size & 1 \\
    Gradient accumulation steps & 8 \\
    Learning rate & 1e-4 \\
    Epoches & 1 \\
    Max length & 2048 \\
    Optimizer & AdamW \\
    Scheduler & cosine \\
    Warmup ratio & 0.1 \\
    LoRA rank & 8 \\
    LoRA alpha & 16 \\
    LoRA dropout & 0 \\
    \bottomrule
    \end{tabular}
  \caption{Hyper-parameters for LM finetuning.}
  \label{tab:hparams}
\end{table}

%% file: figures/promp_declarative.tex
\begin{figure}
\begin{promptbox}{Prompt Template for \textit{Declarative}}
\begin{CJK}{UTF8}{gbsn}
You are a helpful assistant. You are given a query and a target new. Please generate the declaration form of the query and target new.\\

Here is an example:

\textbf{Query:} 2024年华大智造的总资产（亿元）是多少？

\textbf{Target new:} 103.15

\textbf{Declaration:} 2024年华大智造的总资产（亿元）是103.15。\\

Please generate the declaration form of the query and target new below:

\textbf{Query: \{query\}}

\textbf{Target new: \{target\_new\}}

\textbf{Declaration: }
\end{CJK}
\end{promptbox}
\caption{Prompt template for the \textit{declarative} form.}
\label{fig:prompt-declarative}
\end{figure}

%% file: tables/knowedit_stat.tex
\begin{table}[htbp]
\small
\setlength{\tabcolsep}{0.8mm}
  \centering
    \begin{tabular}{lcccc}
    \toprule
    \textbf{Dataset} & ZsRE  & WikiBio & WikiData$_\text{recent}$ & WikiData$_\text{counterfact}$ \\
    \midrule
    \textbf{\# Train} & 10000 & 464   & 570   & 1428 \\
    \textbf{\# Test} & 1304  & 306   & 1266  & 885 \\
    \bottomrule
    \end{tabular}
  \caption{Data statistics of KnowEdit.}
  \label{tab:knowedit-stat}
\end{table}

%% file: figures/prompt_aliased.tex
\begin{figure}
\begin{promptbox}{Prompt Template for \textit{Aliased}}
\begin{CJK}{UTF8}{gbsn}
You are a helpful assistant. You are given a query, a target new, and a declaration. Please generate a paraphrased sentence of the declaration with the central term translated to English.\\

Here is an example:

\textbf{Query:} 2024年华大智造的总资产（亿元）是多少？

\textbf{Target new:} 103.15

\textbf{Declaration:} 2024年华大智造的总资产（亿元）是103.15。

\textbf{Paraphrased sentence:} 2024年华大智造的Total Assets(100 million yuan)是103.15。\\

Here is another example:

\textbf{Query:} 2024年9月中国的订销报纸份数当期值(万份)是多少？

\textbf{Target new:} 137885.2 

\textbf{Declaration:} 2024年9月中国的订销报纸份数当期值(万份)是137885.2。

\textbf{Paraphrased sentence:} 2024年9月中国的Issue of Newspapers, Current Period Value(10000 pieces)是137885.2。\\

Please generate a paraphrased sentence of the declaration below:

\textbf{Query: \{query\}}

\textbf{Target new: \{target\_new\}}

\textbf{Declaration: \{declaration\}}

\textbf{Paraphrased sentence:} 

\end{CJK}
\end{promptbox}
\caption{Prompt template for the \textit{aliased} form.}
\label{fig:prompt-aliased}
\end{figure}

%% file: tables/knowedit_all.tex
\begin{table*}[htbp]
\small
\setlength{\tabcolsep}{1.2mm}
  \centering
    \begin{tabular}{l|l|cccc|ccc|cccc|cccc}
    \toprule
    \multicolumn{1}{c|}{\multirow{2}[0]{*}{\textbf{Model}}} & {\multirow{2}[0]{*}{\textbf{Method}}} & \multicolumn{4}{c|}{\textbf{ZsRE}} & \multicolumn{3}{c|}{\textbf{WikiBio}} & \multicolumn{4}{c|}{\textbf{WikiData$_\text{recent}$}} & \multicolumn{4}{c}{\textbf{WikiData$_\text{counteract}$}} \\         
     && \textbf{ES} & \textbf{L} & \textbf{P} & \textbf{HM} & \textbf{ES} & \textbf{L} & \textbf{HM} & \textbf{ES} & \textbf{L} & \textbf{P} & \textbf{HM} & \textbf{ES} & \textbf{L} & \textbf{P} & \textbf{HM} \\
    \midrule
     \multirow{7}[0]{*}{\textsc{Llama-2}} & SERAC & 94.9  & 44.9  & 40.6  & 52.2  & 98.4  & 70.1  & 81.9  & 92.0  & 33.6  & 47.0  & 48.4  & 78.0  & 29.3  & 42.7  & 42.6 \\
          & IKE   & 99.5  & 46.9  & 67.6  & 65.0  & 99.7  & 41.6  & 58.7  & 85.0  & 45.8  & 57.3  & 58.8  & 85.7  & 51.5  & 60.6  & 63.0 \\
          & EREN  & 36.7  & 67.1  & 47.8  & 47.6  & 60.8  & 64.0  & 62.4  & 44.4  & 59.8  & 38.5  & 46.0  & 25.6  & 67.8  & 25.4  & 32.2 \\
          & WISE  & 71.1  & 67.5  & 51.1  & 61.9  & 95.5  & 99.2  & 97.3  & 80.1  & 48.9  & 53.2  & 58.0  & 72.9  & 56.6  & 50.7  & 58.7 \\
          & RECIPE & 96.4  & 60.0  & 52.6  & 65.1  & 63.8  & 61.9  & 62.8  & 80.8  & 61.2  & 51.4  & 62.2  & 55.6  & 66.8  & 36.8  & 49.9 \\
          & LTE   & \textbf{99.6} & 49.2  & 69.2  & 66.9  & \textbf{100.0} & 47.2  & 64.1  & \textbf{100.0} & 55.0  & 74.6  & 72.1  & \textbf{98.4} & 63.6  & 73.6  & 76.0 \\
          & KEDAS & \textbf{99.6} & \textbf{91.0} & \textbf{73.6} & \textbf{86.7} & \textbf{100.0} & \textbf{100.0} & \textbf{100.0} & \textbf{100.0} & \textbf{77.6} & \textbf{75.9} & \textbf{83.1} & \textbf{98.4} & \textbf{76.3} & \textbf{77.1} & \textbf{82.8} \\
    \midrule
    \multirow{5}[0]{*}{\textsc{Qwen2.5}} & IKE   & 96.7  & 42.2  & 65.1  & 60.7  & \textbf{99.3} & 33.4  & 50.0  & 93.6  & 40.7  & 61.7  & 58.3  & 86.2  & 48.4  & 59.3  & 61.1 \\
          & EREN  & 34.4  & 60.7  & 46.0  & 44.6  & 58.1  & 53.0  & 55.4  & 42.1  & 55.6  & 37.4  & 43.8  & 21.5  & 62.4  & 22.0  & 27.8 \\
          & WISE  & 53.0  & 14.7  & 25.1  & 23.7  & 86.6  & \textbf{100.0} & 92.8  & 54.0  & \textbf{96.8} & 33.0  & 50.7  & 45.4  & 17.9  & 30.3  & 27.1 \\
          & LTE   & \textbf{99.1} & 48.0  & 65.4  & 64.9  & 98.4  & 38.7  & 55.5  & \textbf{99.7} & 42.5  & 68.9  & 62.4  & \textbf{98.0} & 49.4  & 69.9  & 67.0 \\
          & KEDAS & \textbf{99.1} & \textbf{89.6} & \textbf{70.2} & \textbf{84.5} & 98.4  & \textbf{100.0} & \textbf{99.2} & \textbf{99.7} & 67.1  & \textbf{70.5} & \textbf{76.7} & \textbf{98.0} & \textbf{66.6} & \textbf{73.4} & \textbf{77.2} \\
    \bottomrule
    \end{tabular}
  \caption{Main results of sequential editing. Metrics include edit success (ES), locality (L), portability (P) and harmonic mean (HM). The highest scores are \textbf{bolded}.} 
  \label{tab:knowedit-all}
\end{table*}

%% file: tables/knowedit_single.tex
\begin{table*}[htbp]
\small
\setlength{\tabcolsep}{1.5mm}
  \centering
    \begin{tabular}{l|cccc|ccc|cccc|cccc}
    \toprule
      {\multirow{2}{*}{\textbf{Method}}} & \multicolumn{4}{c|}{\textbf{ZsRE}} & \multicolumn{3}{c|}{\textbf{WikiBio}} & \multicolumn{4}{c|}{\textbf{WikiData$_\text{recent}$}} & \multicolumn{4}{c}{\textbf{WikiData$_\text{counteract}$}} \\
       
     & \textbf{ES} & \textbf{L} & \textbf{P} & \textbf{HM} & \textbf{ES} & \textbf{L} & \textbf{HM} & \textbf{ES} & \textbf{L} & \textbf{P} & \textbf{HM} & \textbf{ES} & \textbf{L} & \textbf{P} & \textbf{HM} \\
    \midrule
        LoRA  & \textbf{100.0} & 75.7  & 56.5  & 73.3  & \textbf{100.0} & 81.7  & 89.9  & \textbf{100.0} & 56.3  & 64.5  & 69.3  & \textbf{100.0} & 70.7  & 68.9  & 77.6 \\
          MEND  & 97.9  & 96.4  & 61.1  & 81.2  & 58.3  & 90.8  & 71.0  & 94.4  & 77.9  & 43.3  & 64.5  & -     & -     & -     & - \\
          ROME  & 96.7  & 53.7  & 52.7  & 62.5  & 96.0  & 62.7  & 75.8  & 97.0  & 54.7  & 55.7  & 64.5  & 98.5  & 50.4  & 55.8  & 62.6 \\
          SERAC & 97.0  & 60.0  & 61.3  & 69.3  & 98.4  & 72.8  & 83.7  & 99.0  & 36.4  & 64.9  & 56.6  & 99.0  & 32.4  & 76.7  & 55.6 \\
          IKE   & 99.8  & 47.1  & 78.1  & 68.1  & 99.7  & 41.3  & 58.4  & 85.0  & 45.9  & 62.6  & 60.6  & 87.3  & 52.0  & 72.2  & 67.4 \\
          EREN  & 36.7  & 67.1  & 47.8  & 47.6  & 60.8  & 64.0  & 62.4  & 44.4  & 59.8  & 38.5  & 46.0  & 25.6  & 67.8  & 25.4  & 32.2 \\
          WISE  & 99.9  & \textbf{99.7} & 47.4  & 72.9  & \textbf{100.0} & \textbf{100.0} & \textbf{100.0} & 98.3  & 82.7  & 62.2  & 78.2  & 99.6  & 82.3  & 71.1  & 82.8 \\
          AlphaEdit & 97.5  & 66.9  & 51.2  & 67.0  & 99.0  & 71.8  & 83.2  & 97.8  & 60.8  & 52.3  & 65.5  & 98.9  & 64.1  & 47.7  & 64.3 \\
          RECIPE & 96.6  & 78.3  & 55.9  & 73.1  & 63.8  & 94.6  & 76.2  & 81.0  & 71.1  & 53.2  & 66.3  & 55.7  & 76.9  & 38.3  & 52.6 \\
          LTE   & 99.9  & 49.2  & \textbf{77.6} & 69.4  & \textbf{100.0} & 47.2  & 64.1  & \textbf{100.0} & 55.0  & \textbf{82.0} & 74.3  & \textbf{100.0} & 65.0  & \textbf{88.0} & 81.6 \\
          KEDAS & 99.6  & 96.3  & 77.3  & \textbf{89.9} & \textbf{100.0} & \textbf{100.0} & \textbf{100.0} & \textbf{100.0} & \textbf{98.1} & 81.1  & \textbf{92.2} & 98.4  & \textbf{99.4} & 85.9  & \textbf{94.1} \\
    \bottomrule
    \end{tabular}
  
  \caption{Results of single editing on KnowEdit with Llama-2-7B-Chat. The highest scores are \textbf{bolded}. ``-'' means the method encounters out-of-memory (OOM) or tensor overflow on our GPU.}
  \label{tab:knowedit-single}
\end{table*}